\documentclass[preprint,12pt,authoryear]{elsarticle}

\usepackage{amssymb}
\usepackage{amsmath}

\usepackage{tabularx}
\usepackage{natbib}
\usepackage{booktabs}
\usepackage{multirow}
\setcitestyle{numbers,square}

\journal{Nuclear Physics B}

\begin{document}

\begin{frontmatter}



\title{Incomplete Multi-Label Image Recognition by Co-learning Semantic-Aware Features and Label Recovery}


\author[1]{Zhi-Fen He}
\ead{zfhe323@nchu.edu.cn}

\author[1]{Ren-Dong Xie}
\ead{2307070100001@stu.nchu.edu.cn}

\author[1]{Bo Li \corref{cor1}}
\ead{libo@nchu.edu.cn}

\author[1]{Bin Liu}
\ead{nyliubin@nchu.edu.cn}

\author[1]{Jin-Yan Hu}
\ead{2407070100102@stu.nchu.edu.cn}

\address[1]{School of Mathematics and Information Science, Nanchang Hangkong University, Nanchang, 330063, China}

\cortext[cor1]{Corresponding author: Tel.: +86 0791-83863755; Postal address: No.696, Fenghe South Road, Honggutan District, Nanchang, Jiangxi
330063, China.}

\begin{abstract}
Multi-label image recognition with incomplete labels is a challenging yet vital task in computer vision, which faces two fundamental challenges: learning semantic-aware features and recovering missing labels. In this paper, we propose a Co-learning framework for Semantic-aware features and Label recovery (CSL), designed to address both challenges in a unified learning paradigm. Specifically, we develop a semantic-related feature learning module that captures robust semantic-related representations by discovering semantic information and label correlations. Furthermore, a semantic-guided feature enhancement module is introduced to generate highly discriminative semantic-aware features by effectively aligning visual and semantic spaces. Finally, we present a collaborative learning framework that integrates semantic-aware feature learning with label recovery. This framework not only dynamically enhances the discriminability of semantic-aware features but also adaptively infers and recovers missing labels, thereby forming a mutually reinforcing mechanism between the two processes. Extensive experiments on three widely used public datasets (MS-COCO, VOC2007, and NUS-WIDE) demonstrate that CSL outperforms state-of-the-art methods for incomplete multi-label image recognition.
\end{abstract}




\end{frontmatter}


\section*{Code Metadata}
The code is available on GitHub: https://github.com/Qingdandian/CLSL

\section{Introduction}
\label{}

Multi-Label Image Recognition (MLIR), which aims to assign multiple semantic labels to a single image, is widely applied in fields such as medical imaging analysis \cite{re1}, visual retrieval systems \cite{re2}, and scene understanding \cite{re3}. Most existing methods typically assume that the labels are fully annotated \cite{re17,re18,re19,re20,re21,re22,re23,re24,re25}. However, in practical applications, obtaining large-scale multi-label datasets with complete annotations is extremely time-consuming and labor-intensive. Therefore, recent studies have tended to focus on the task of multi-label image recognition with incomplete labels, where only a small number of positive and negative labels are known, while the remaining labels are unknown (see Figure 1). This task has been widely applied because it does not require complete multi-label annotations to be provided for each image.

A simple and straightforward approach is to treat unknown labels as negative ones or directly ignore them \cite{re4,re5, re6}, before applying traditional MLIR methods to the task. However, this strategy often leads to suboptimal performance, as it either introduces incorrect annotations by treating unknown labels as negative or ignores some potentially useful annotation information. For instance, Durand et al. \cite{re6} introduced a partial binary cross-entropy (partial-BCE) loss that simply ignores unknown labels. To address this issue, some researchers \cite{re6,re7,re8,re9,re10} have shifted their focus to leveraging known labels for recovering missing labels. For example, Chen et al. \cite{re7,re8} adopted semantic transfer or prototype blending, and attempted to model cross-modal correlations to recover incomplete labels. Nevertheless, these methods heavily rely on prior assumptions, and lack the capability to extract high-quality semantic-aware features when facing extreme annotation sparsity. Additionally, they often overlook discriminative fine-grained visual cues, which further restricts their robustness and generalization ability.

In addition, vision-Language Pre-training (VLP)-based methods, such as CLIP \cite{re11}, have emerged as a powerful paradigm in recent years and are widely adopted to tackle the incomplete multi-label image recognition problem \cite{re12, re13, re14,re15,re16,re31}. Among these VLP-based approaches, some designs specialize in prompt templates \cite{re12} or integrate structured semantic priors to model label correlations \cite{re31}, while others leverage textual captions as images to augment training \cite{re14,re15}. Although these VLP-based methods have achieved acceptable performance, they still exhibit notable limitations: 1) Current methods \cite{re12,re13,re31} fail to effectively utilize label information. This not only leads to discard valuable information implicitly associated with unknown labels, but also impairs the model’s ability to learn accurate label correlations between labels. 2) Existing methods \cite{re14,re15} tend to prioritize global semantic alignment and struggle to capture fine-grained spatial structures, leading to underutilize local visual cues that are critical for distinguishing between multiple labels, especially in scenarios with extremely sparse annotations. While synthetic data generation \cite{re16} has been explored to bridge the modality gap between vision and language, it often introduces domain shift, further compromising performance.

\begin{figure}
\centerline{\includegraphics[width=1\textwidth]{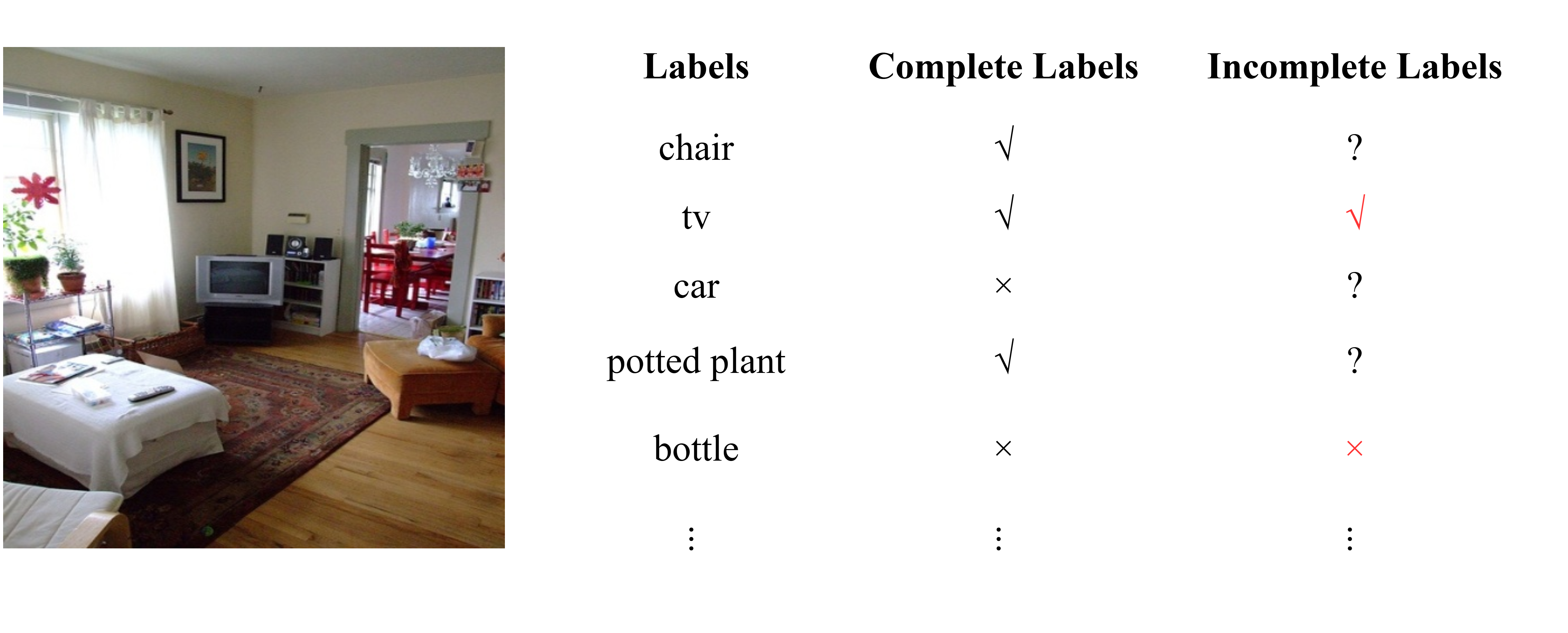}}
\caption{Examples of fully and partially labeled image annotations. Here, "$\checkmark$" indicates positive labels, "$\times$" corresponds to negative labels, and "$?$" refers to missing or unannotated labels.}
\label{Fig.1}
\end{figure}

Based on the above analysis, we propose a \textbf{C}o-learning \textbf{S}emantic-Aware Features and \textbf{L}abel Recovery (CSL) framework to address the challenge of multi-label image recognition with incomplete labels. First, to effectively capture semantic information and label correlations, a semantic-related feature learning module is designed by fusing global visual features with label embeddings to generate semantic-related features. Next, to enhance the alignment between visual and semantic spaces and further boost the discriminative capability of the fused features, a low-rank bilinear model is utilized to produce high-quality semantic-aware features. Finally, a collaborative learning framework for semantic-aware feature learning and label recovery is introduced, which adaptively infers and recovers missing labels, treating the recovered labels as pseudo-labels. These pseudo-labels then iteratively guide the optimization of the semantic-aware feature learning process, forming a mutually reinforcing cycle that continuously refines both features quality and label completeness. 

The main contributions are summarized as follows:

1) We propose a semantic-aware feature learning framework for incomplete multi-label image recognition. Comprising a semantic-related feature learning module and a semantic-guided feature enhancement module, this framework effectively models visual-semantic interactions and generates high-quality discriminative semantic-aware features.

2) We design a collaborative learning strategy that enables the joint optimization of semantic-aware feature learning and label recovery. This strategy can not only dynamically enhance the discriminative capability of visual features but also adaptively recover missing labels.

3) Extensive experiments on three widely used multi-label benchmarks (MS-COCO, VOC2007, and NUS-WIDE) demonstrate that our method achieves state-of-the-art (SOTA) performance for multi-label image recognition with incomplete labels.



\section{Related Work}
\subsection{Multi‐Label Image Recognition}
Multi-label image recognition (MLIR) is a core and indispensable task in computer vision. Early approaches typically formulated MLIR as a set of independent binary classification problems, where a binary classifier is trained for each label. However, these methods ignore label dependencies and thus limit recognition performance. Recent work has focused on explicitly or implicitly learning label correlations by modeling label sequences with Recurrent Neural Networks (RNNs) \cite{re17}. For example, Wang et al. \cite{re17} proposed a unified CNN–RNN framework that uses a CNN to extract image features and an RNN (LSTM) to generate multiple labels in a sequential manner, thereby explicitly modeling label dependencies to improve multi-label image recognition performance. Some research captures graph-based label representation by Graph Convolutional Networks (GCNs)\cite{re18,re19,re20} or knowledge-statistical (KS) graphs \cite{re21}. For instance, Chen et al. \cite{re19} formalized the label space as a graph and then employed GCNs to generate semantic-aware label embeddings. Wang et al. \cite{re21} constructed a label graph on image feature/region graphs to fuse semantic label relationships with spatial/feature information. There are also works focused on the attention mechanism or object localization \cite{re22,re23,re24,re25}. For example, Ye et al. \cite{re22} introduced an attention-based dynamic GCN framework that constructs and updates connections between labels and image regions. Zhu et al. \cite{re24} incorporated residual attention modules into the backbone to emphasize critical channels/regions while preserving training stability. Zhu et al. \cite{re25} leveraged semantic guidance to enhance visual representations, then fusing semantic cues with image features. Additionally, Vision-Language Pre-training (VLP)-Based methods have emerged as a powerful paradigm in recent years. Benefiting from large-scale pre-trained models aligned with image-text data, these methods leverage specially designed prompts or lightweight adapters to transfer pre-trained cross-modal representations to the MLIR task effectively, achieving state-of-the-art robust multi-label image recognition performance \cite{re26, re27, re28}. Despite the remarkable progress made by the aforementioned methods, they heavily rely on large-scale fully annotated image datasets for training, which limits their applicability and performance in label-limited scenarios and remains a major challenge in practical MLIR applications. 

\subsection{Incomplete Multi‐Label Image Recognition}
In recent years, multi-label image recognition with incomplete labels has attracted significant attention, and a series of methods have been proposed. A naïve approach is to treat unannotated labels as negative ones \cite{re4,re5}. However, this strategy often incorrectly classifies actual positive labels as negative ones, resulting in suboptimal performance. To address this issue, some work focuses on utilizing only positive labels for model training. For instance, Durand et al. \cite{re6} proposed a partial-label multi-label classification method that trains a deep convolutional network (partial-BCE). This method exploits only known labels and directly ignores unannotated labels. Similarly, Yuan et al. \cite{re29} developed a Positive-Unlabeled learning approach (PU-MLC), which trains the multi-label classification model using only positive labels and unlabeled data, avoiding the risk of mislabeling unannotated positive labels as negative. Ben-Baruch et al. \cite{re9} introduce a class-aware selective loss for partial-annotation settings, selectively computing and weighting training signals per class to suppress noise from unannotated or unreliable labels. Beyond relying solely on positive labels, some researchers have attempted to leverage semantic or visual correlations to recover incomplete annotations. For example, Chen et al. \cite{re7} propose a structured semantic transfer approach that leverages semantic relationships between labels to propagate knowledge from well-annotated to partially or unannotated samples, using semantic constraints and structured propagation to mitigate the effects of missing labels. Pu et al.\cite{re8} designed a unified semantic-aware representation blending (SARB) framework, which takes advantage of semantic representations at both the instance and prototype level to recover unannotated labels. Ma et al. \cite{re30} proposed a multimodal category prototype that exploits intra-category and inter-category semantic relationships to estimate unknown labels, thereby generating effective pseudo-labels. However, these methods often conduct feature learning and label recovery in a stepwise manner, thus failing to leverage the mutual promotion between the two processes. 

In addition, CLIP-based vision language pretraining methods have been widely adopted in multi-label settings with limited annotations.
For example, Sun et al. \cite{re12} encoded positive and negative contexts with class names as part of the linguistic input (i.e., prompts), which can quickly adapt to multi-label recognition tasks with limited annotations. Ding et al. \cite{re31} proposed the SCP-Net network, which aims to mine structured semantic relationships to recover unknown labels. Guo et al. \cite{re14} treated texts as images for prompt tuning and introduced double-grained prompt tuning (TaI-DPT) to extract coarse- and fine-grained embeddings. Feng et al. \cite{re16} leveraged the pre-trained text-to-image generation model to generate realistic and diverse images from text captions, and then aligned the generated images with their corresponding text, effectively reducing the modality gap inherent in text-as-image models \cite{re14}. Despite these advances, most CLIP-based methods depend mainly on global alignment through cosine similarity, and thus underusing fine-grained local visual cues that are essential for recovering unknown multi-label annotations.

\section{The Proposed Method}

\begin{figure}[!htbp]  
    \centering
    \includegraphics[width=1.0\textwidth]{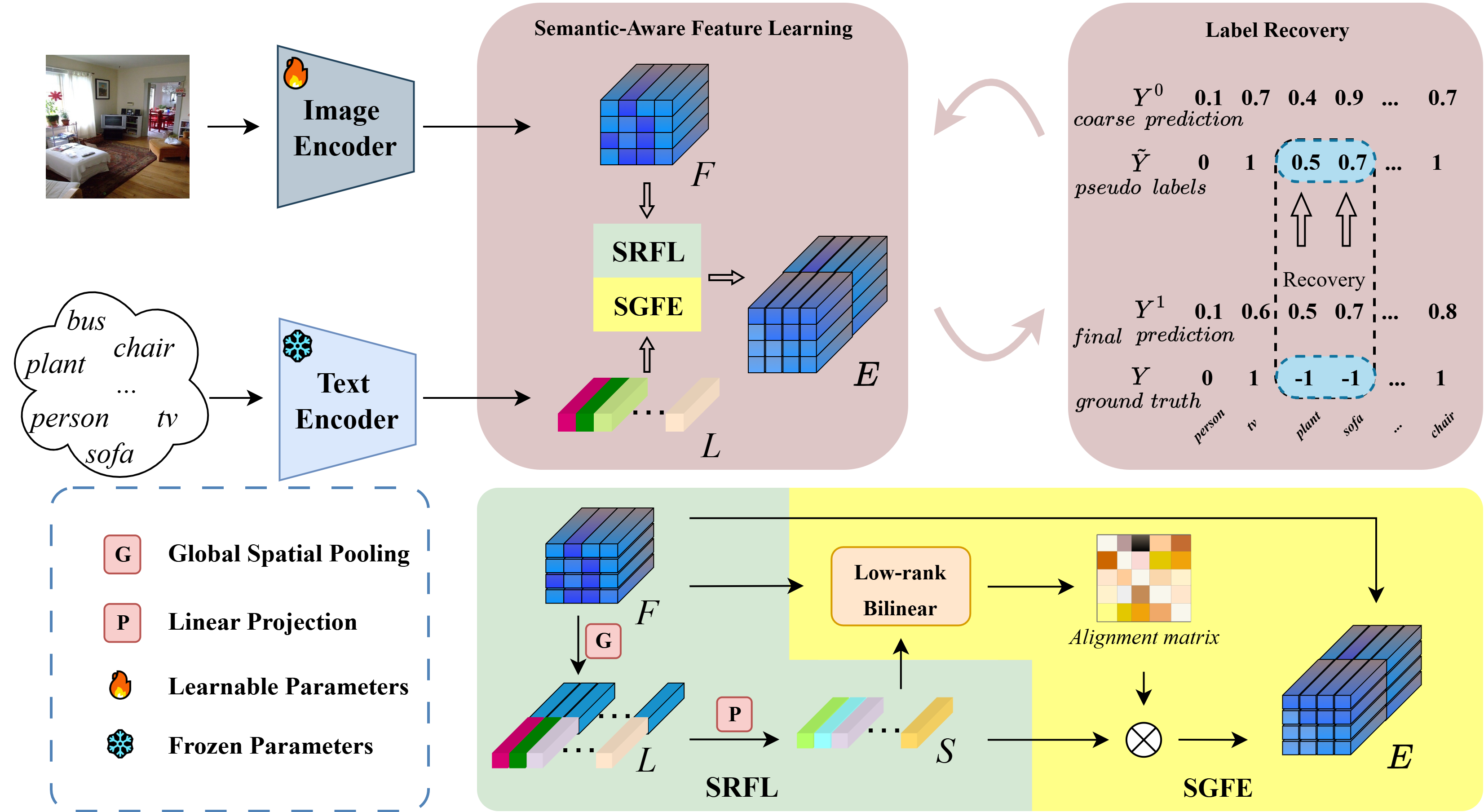}  
    \caption{Pipeline of the CSL Method.}
    \label{Fig.2}
\end{figure}

\subsection{Problem Definition and Overview}
Incomplete multi-label image recognition aims to accurately predict all relevant semantic labels in an image when only a partial set of labels is available during training. Let $D=\{(X_i,Y_i)\}_{i=1}^N$ denote the training dataset, where $N$ represents the number of images. Specifically, $X_i \in R^{H  \times W  \times 3}$ stands for the $i$-th image, while $Y_i = [y_{i1},...,y_{iC}] \in \{-1, 0, 1\}^C$ is the corresponding label vector, where $C$ denoting the total number of labels. For a given label $j$, $y_{ij}=1$ indicates that label $j$ is present in the $i$-th image, $y_{ij}=0$ indicates that label $j$ is absent, and $y_{ij}=-1$ denotes that the presence of label $j$ is unknown.


Existing approaches often struggle to jointly learn semantically meaningful features and accurately recover missing labels, particularly under severe annotation sparsity. Many methods either treat unannotated labels as negative instances or rely on globally pooled features that lack local discriminability and semantic sensitivity. To address these limitations, we propose a collaborative learning framework that unifies semantic-aware feature learning with dynamic label recovery.

As illustrated in Figure 2, The proposed CSL framework consists of two core modules: (1) \textbf{Semantic-Aware Feature Learning}. This module extracts highly discriminative semantic-aware features under severely incomplete labels. It comprises two sub-components: a Semantic-Related Feature Learning (SRFL) component, which encodes semantic correlations into visual representation features, and a Semantic-Guided Feature Enhancement (SGFE) component, which integrates image features with semantic-related features via a low-rank bilinear model to produce refined semantic-aware features. (2) \textbf{Label Recovery}. This module leverages the discriminative semantic-aware features to accurately recover missing annotations. These two modules are jointly optimized under a collaborative learning strategy, forming a close-loop system where label recovery and semantic-aware feature learning mutually reinforce each other. This integrated design not only enhances feature discriminability but also significantly improves multi-label recognition performance in incomplete label scenarios.

\subsection{Semantic-Aware Feature Learning}
The Semantic-Aware Feature Learning module is designed to extract and enhance visual features that are semantically aligned with label information, even under incomplete annotation. It consists of two complementary components: Semantic-Related Feature Learning and Semantic-Guided Feature Enhancement, which work collaboratively to enable robust feature representation learning in incomplete label scenarios.

The process begins by encoding an input image $X_i$ using a visual backbone (e.g., ResNet-101 \cite{re32}) to produce a spatial feature map. To capture global contextual information, the feature map is flatten and processed through self-attention layers based on  a Transformer architecture \cite{re33}, yielding structured image features $F_i\in R^{P \times d_v}$, where $P = H \times W$ denotes the number of patches and $d_v$ is the feature dimension. (The subscript $i$ is omitted hereafter for notational simplicity.) In parallel, each label is embedded into a semantic vector via a text encoder, forming a label embedding set $L=\{l_1, l_2, \ldots, l_C\}$, where $l_j \in R^{d_t}$ for $1 \leq j \leq C$, and $d_t$ denotes the label embedding dimension. 

\textbf{Semantic-Related Feature Learning.}
To effectively capture multi-label semantic information and inherent label correlations under incomplete supervision, we propose the Semantic-Related Feature Learning (SRFL) module.The SRFL is designed to extract discriminative semantic-related features, even when dealing with severely incomplete annotations. Specifically, a global feature vector $F^G \in R^{d_v}$ is first obtained via global spatial pooling, and then fused with the label embeddings $L$ to enhance semantic consistency. The resulting semantic-related features $S \in R^{C \times d_v}$ are formulated as follows:
\begin{equation}
    F^{\mathrm{G}}=\mathrm{GSP}(F),
\end{equation}
\begin{equation}
    S=\text { Linear }\left(\operatorname{concat}\left(F^{\mathrm{G}}, L\right)\right),
\end{equation}
where GSP($\cdot$) denotes global spatial pooling, and Linear($\cdot$) represents a linear projection.

\textbf{Semantic-Guided Feature Enhancement.} 
To enhance the alignment between visual and semantic representations, we propose the Semantic-Guided Feature Enhancement (SGFE) module based on a low-rank bilinear pooling model. The SGFE employs a semantic-aware attention mechanism to effectively integrate image features $F=\left\{f_{p}|_{p=1}^{P}\right\} \in R^{P \times d_{\mathrm{v}}}$ with semantic-related features $S=\left\{\left.s_{c}\right|_{c=1} ^{C}\right\} \in R^{C \times d_{\mathrm{v}}}$. Concretely, the low-rank bilinear model~\cite{re34} first constructs a cross-modal alignment matrix, which is then utilized to adaptively weight and fuse the semantic and visual representations, producing discriminative and robust semantic-aware features $E$. This process effectively improves the performance of multi-label recognition, even under severely incomplete annotation. The resulting semantic-aware features $E=\left\{\left.e_{p}\right|_{p=1} ^{P}\right\} \in R^{P \times 2d_{v}}$ are formulated as follows:
\begin{equation}
    e_{p}=\operatorname{concat}\left(\sum_{c=1}^{C} b_{p c} s_{c}, f_{p}\right),
\end{equation}
where $b_{p c}$ is computed as follows:
\begin{equation}
    b_{p c}=\frac{\exp \left(a_{p, c} / t\right)}{\sum_{c^{\prime}} \exp \left(a_{p, c^{\prime}} / t\right)}.
\end{equation}
where $t$ is a temperature coefficient used to control the smoothness of the attention distribution, $a_{p,c}$ represents the attention weight between the $p$-th image patch and the $c$-th label, and $\mathrm{A}=\{\left.a_{p,c}\right \}$ denotes the attention matrix which is computed by:
\begin{equation}
    \mathrm{A}=w((\tanh ((F U) \odot(S V))) P+b),
\end{equation}
where $P \in R^{d_1 \times d_2}$, $U \in R^{d_v \times d_1}$, $V \in R^{d_v \times d_1}$ and $b \in R^{d_2}$ are learnable parameters, $d_1$ and $d_2$ denote the dimensions of the joint embedding space and the output features, tanh($\cdot$) is the hyperbolic tangent function, $'\odot'$ denotes element‐wise multiplication, and $w(\cdot)$ is a linear transformation mapping from $R^{d_2}$ to $R$.  

\subsection{Label Recovery}
In incomplete multi-label image recognition, many existing methods directly discard missing labels and rely solely on the observed ones, resulting in sub-optimal model performance. To overcome this limitation, we introduce a joint learning method that simultaneously addresses multi-label image classification and label recovery.

Given the semantic‐aware features $E \in R^{P \times 2d_v}$, we first perform location‐wise classification using a classifier  $\mathrm{CLS_1} \in R^{2d_v \times C}$, producing a per‐location classification matrix $M \in R^{P \times C}$ as follows: 
\begin{equation}
    M=E \times \mathrm{CLS}_{1}.
\end{equation}

Next, softmax normalization is applied to $M$ along the spatial dimension, followed by a weighted aggregation of region scores to obtain the final multi-label prediction scores $Y^1 \in R^C$:  
\begin{equation}
    Y^{1}=\sum_{p} M \odot \operatorname{softmax}_{P}(M),
\end{equation}
where $\operatorname{softmax}_{P}(\cdot)$ denotes softmax operation applied across the patch dimension, and $'\odot'$ represents element‑wise multiplication. 

Subsequently, the refined prediction $Y^1$ is used to complete the missing entries in the ground‑truth label vector $Y$, forming a pseudo‑label matrix $\tilde{Y} \in R^{C}$. Specifically, the positions originally missing in $Y$ are filled with the corresponding predicted probabilities from $Y^1$, while the annotated labels remain unchanged (i.e., preserved as 0 or 1). 
\begin{equation}
    \tilde{Y}_{i j}=\left\{\begin{array}{lll}
1, & \text { if } & y_{i j}=1 \\
Y_{i j}^{1}, & \text { if } & y_{i j}=-1 \\
0, & \text { if } & y_{i j}=0
\end{array}\right.
\end{equation}
where $y_{ij}$ denotes the $j$-th label of the $i$-th image.

\subsection{Collaborative Learning and Inference}
To address the key challenges in incomplete multi-label image recognition: the limited discriminability of learned feature representations and the inaccuracy in recovering missing labels, we introduce a collaborative learning framework that jointly optimizes semantic-aware feature learning and label recovery.

First, the initial image features $F \in R^{P \times d_v}$ are processed through the classifier $\mathrm{CLS}_0 \in R^{d_v \times C}$ and then aggregated via global max pooling, producing a coarse prediction score vector $Y^0 \in R^C$. Notably, $Y^0$ captures the most salient and readily activated labels in the image, representing a preliminary, global-level assessment of label presence. To strengthen learning under label incompleteness, we supervise $Y^0$ using the pseudo-labels $\tilde{Y}$. Meanwhile, to mitigate potential noise amplification, the refined prediction $Y^1$ is supervised only with the original known ground-truth labels. Finally, with $Y^0$, $Y^1$ and $\tilde{Y}$ obtained from the preceding stages, we optimize the model jointly using the $\mathrm{ASL}$ loss \cite{re35}:
\begin{equation}
    {L_\mathrm{A S L}(p,y)}=\left\{\begin{array}{ll}
(1-p)^{\gamma+} \log (p), & y^{j}=1 \\
\left(p_{m}\right)^{\gamma-} \log \left(1-p_{m}\right), & y^{j}=0
\end{array}\right.
\end{equation}
where $y$ denotes the ground truth value, $p$ denotes the predicted value, $p_m=\mathrm{max}(p-c, 0)$ represents the shifted probability for hard negative samples, and $\gamma+$, $\gamma-$ and $c$ are hyperparameters. 

The overall loss function is formulated as follows:
\begin{equation}
    L=\lambda_{1}{L_\mathrm{ A S L}}(Y^{1}, Y)+\lambda_{2} {L_\mathrm{A S L}}(Y^{0}, \tilde{Y}),
\end{equation}
where $\lambda_1$ and $\lambda_2$ are balancing hyperparameters. The first term constrains the refined predictions  $Y^1$ using the original ground-truth labels $Y$, while the second term enables self-correction by aligning the coarse predictions $Y^0$ with the reconstructed pseudo-labels $\tilde{Y}$. By integrating label recovery and semantic feature learning into a collaborative loop, our approach enhances both the discriminability of visual representations and the accuracy of the recovered label set. The entire framework is trained end-to-end using the $\mathrm{ASL}$\cite{re35} loss, which effectively leverages both original and recovered annotations.

During inference, the model relies solely on the refined predictions $Y^{1}$, enabling effective multi-label image recognition under incomplete supervision.

\section{Experiments}

\subsection{Experimental Settings}
\textbf{Datasets.} We conduct experiments on three popular benchmark datasets, including MS-COCO~\cite{re36}, PASCAL VOC 2007 (VOC2007)~\cite{re37} and NUS-WIDE~\cite{re38}, to evaluate multi-label image recognition with incomplete labels. Detailed dataset statistics are provided in Table 1, where Label Cardinality stands for the average of labels per image. To create the training set with incomplete labels, we randomly mask out some labels from the fully labeled training set, and use the remaining labels for training as in previous work~\cite{re12}. In this paper, the proportion of masked labels varies from 90$\%$ to 10$\%$, resulting in 10$\%$ to 90$\%$ known labels.
\begin{table}[h]
    \centering
    \footnotesize
    {\textbf{Table 1:} Statistics of Multi-label Image Datasets. }\\[0.5em]  
    \label{Table.1}
    \begin{tabular}{lcccccc}
        \toprule
        Dataset & Train & Test & Label & \multicolumn{3}{c}{Label Cardinality} \\
        \midrule
        MS-COCO & 82783 & 40504 & 80 & \multicolumn{3}{c}{2.9}\\
        VOC2007 & 5011 & 4952 & 20 & \multicolumn{3}{c}{1.5} \\
        NUS-WIDE& 125449& 83898& 81& \multicolumn{3}{c}{2.4} \\
        \bottomrule
    \end{tabular}
\end{table}

\textbf{Evaluation criteria.}
We utilize widely adopted metric: mean Average Precision (mAP) over all labels, which is defined as follows: 
\begin{equation}
    \mathrm{mAP}=\frac{1}{C} \sum_{i=1}^{C} \mathrm{AP}(i)
\end{equation}
Where the average precision (AP) for the $i$-th label is defined as 
\begin{equation}
    \mathrm{AP}(i)=\frac{\sum_{k=1}^{R} \widehat{P}_{i}(k) \times r_{i}(k)}{\sum_{k=1}^{R} r_{i}(k)},
\end{equation}
$C$ denotes the number of labels, $r_i(k)$ is the recall of label $i$ at the $k$-th ranked prediction among all predictions, and $\widehat{P}_{i}(k)$ represents the precision of label $i$ at the $k$-th ranked prediction. 

\textbf{Implementation Details.}
We implement the proposed CSL framework in PyTorch using one NVIDIA Tesla A40 GPU. ResNet-101~\cite{re32} is adopted as the image encoder and BERT~\cite{re39} as the text encoder. We use the AdamW optimizer for training and apply RandAugment and Cutout \cite{re27} for data augmentation. To improve training stability, we employ an Exponential Moving Average (EMA) with a decay rate of 0.9997. The asymmetric loss parameters are fixed as $\gamma^+ = 0$ and $\gamma^- = 2$ across all experiments. The detailed experimental parameters for each dataset are provided in Table 2.

\begin{table}[h]
    \centering
    \footnotesize
    \setlength{\tabcolsep}{4pt}  
    {\textbf{Table 2:} Configuration of Experimental Parameters. Note that “vanilla” refers to the image encoder pre-trained on ImageNet, whereas “clip” denotes the image encoder that adopts parameters derived from CLIP.}\\[0.5em]  
    \label{Table.2}
    \begin{tabular}{ l l l l l l l }
        \toprule
        Dataset & Backbone & Input size & Batch size & Learning rate & $\lambda_1$ & $\lambda_2$ \\ 
        \midrule
        MS-COCO & vanilla & 448×448 & 52 & 5e-5 & 1 & 0.8 \\ 
        VOC2007 & vanilla & 448×448 & 64 & 5e-5 & 1 & 0.1 \\ 
        NUS-WIDE & vanilla & 224×224 & 128 & 9e-5 & 1 & 2 \\ 
        \midrule
        MS-COCO & clip & 448×448 & 32 & 5e-5 & 1 & 0.8 \\ 
        VOC2007 & clip & 448×448 & 32 & 2.4e-5 & 1 & 0.1 \\ 
        NUS-WIDE & clip & 224×224 & 128 & 9e-5 & 1 & 2 \\ 
        \bottomrule
    \end{tabular}
\end{table}

\textbf{Baselines.}
To verify the effectiveness of CSL, we compare with the following baselines: 

\textbullet Partial BCE~\cite{re6}, a classification loss that only exploits the proportion of known labels per example. 

\textbullet SSGRL\cite{re20}, a semantic-specific graph representation learning framework comprising a semantic decoupling module and a semantic interaction module. 

\textbullet P-GCN\cite{re19}, an approach that decomposes the visual representation into a set of label-aware features, followed by encoding to generate interdependent image-level prediction scores. 

\textbullet ASL\cite{re35}, a loss function designed to tackle the positive-negative imbalance. 

\textbullet SST\cite{re7}, a method that consists of two complementary transfer modules, which leverages intra-image and inter-image semantic correlations to transfer knowledge from known labels and further generate pseudo-labels of unknown instances. 

\textbullet SARB\cite{re8}, a unified semantic-aware representation blending framework that leverages both instance-level and prototype-level semantic representations to enhance the labeling of unknown instances. 

\textbullet PU-MLC\cite{re29}, a method that directly discards negative labels and employs Positive-Unlabeled (PU) Learning. 

\textbullet DualCoOp\cite{re12}, a CLIP-based method that encodes both positive and negative prompts per label. 

\textbullet DualCoOp++\cite{re18}, a CLIP-based method that independently encodes evidential, positive, and negative prompts per label.

\textbullet SCPNet\cite{re31}, a semantic correspondence prompt network that can thoroughly explore structured semantic priors. 

\textbullet TaI-DPT\cite{re14}, a double-grained prompt tuning method that extracts both coarse-grained and fine-grained embeddings to enhance multi-label recognition performance. 

\textbullet TRM-ML\cite{re30}, a multi-modal method that incorporates category-aware region learning and contrastive learning to improve region-level matching in multi-label recognition tasks. 

\textbullet T2I-PAL\cite{re16}, a text-to-image generation method that narrows modality gaps by using synthesized images and boosts multi-label recognition using class-wise heatmaps and learnable prototypes.

\subsection{Comparison with State-of-the art (SOTA) Methods}
\textbf{Performance on MS-COCO.} We first report the performance comparisons on MS-COCO in Table 3. Table 3 demonstrates that our method outperforms existing SOTA methods on the MS-COCO dataset under the same pre-trained backbone and input resolution setting. Specifically, our method achieves the best overall performance. When compared with ImageNet-pretrained baseline approaches, it yields an average mAP ranging from 1.9$\%$ to 9$\%$. The large performance gap (especially the 9$\%$ gain over certain baselines) highlights the limitations of traditional ImageNet pre-training in capturing label-specific features for MS-COCO, while verifying the effectiveness of our proposed feature enhancement module. When benchmarked against CLIP-based SOTA methods, such as DualCoOp, TRM-ML and T2I-PAL, our proposed method still maintains notable advantages, with its average mAP outperforming these SOTA approaches by 1$\%$ to 7.5$\%$. This result indicates that our framework can achieve more stable and accurate weakly supervised recognition on MS-COCO.
\begin{table}[htbp]
    \centering
    \footnotesize
    {\textbf{Table 3:} Performance comparisons (in mAP, $\%$) of different methods on MS-COCO under varying known label ratios $p$. The best performance in each comparison group is highlighted in \textbf{bold}. Note that “vanilla” refers to the image encoder pre-trained on ImageNet, whereas “clip” denotes the image encoder that adopts parameters derived from CLIP.} \\[0.5em]
    \label{Table.3}
    \resizebox{\textwidth}{!}{%
    \begin{tabular}{l l l ccccccccc c}
        \toprule
        Method & Venue & Backbone & $p{=}0.1$ & $p{=}0.2$ & $p{=}0.3$ & $p{=}0.4$ & $p{=}0.5$ & $p{=}0.6$ & $p{=}0.7$ & $p{=}0.8$ & $p{=}0.9$ & Avg \\
        \midrule
        Partial BCE\cite{re6}           & CVPR$'$19 & vanilla & 61.6 & 70.5 & 74.1 & 76.3 & 77.2 & 77.7 & 78.2 & 78.4 & 78.5 & 74.7 \\
        SSGRL\cite{re20}                 & ICCV$'$19 & vanilla & 62.5 & 70.5 & 73.2 & 74.5 & 76.3 & 76.5 & 77.1 & 77.9 & 78.4 & 74.1 \\
        P-GCN\cite{re19}                 & TPAMI$'$21 & vanilla & 68.3 & 72.3 & 74.4 & 76.3 & 77.9 & 78.6 & 79.7 & 80.7 & 81.6 & 76.6 \\
        ASL\cite{re35}                   & ICCV$'$21 & vanilla & 69.7 & 74.0 & 75.1 & 76.8 & 77.5 & 78.1 & 78.7 & 79.1 & 79.7 & 76.5 \\
        SST\cite{re7}                   & AAAI$'$22 & vanilla & 68.1 & 73.5 & 75.9 & 77.3 & 78.1 & 78.9 & 79.2 & 79.6 & 79.9 & 76.7 \\
        SARB\cite{re8}                  & AAAI$'$22 & vanilla & 71.2 & 75.0 & 77.1 & 78.3 & 78.9 & 79.6 & 79.8 & 80.5 & 80.5 & 77.9 \\
        PU-MLC\cite{re29}                 & ICME$'$24 & vanilla & 75.7 & 78.6 & 80.2 & 81.3 & 82.0 & 82.6 & 83.0 & 83.5 & 83.8 & 81.2 \\
        CSL                  & Ours & vanilla & \textbf{78.2} & \textbf{80.6} & \textbf{82.3} & \textbf{83.1}& \textbf{83.8} & \textbf{84.4}& \textbf{84.7} & \textbf{85.0}& \textbf{85.4} & \textbf{83.1}\\
        \midrule
        SST*\cite{re7}                  & AAAI$'$22 & clip & 69.1 & 78.5 & 79.3 & 79.9 & 80.1 & 80.5 & 81.1 & 80.7 & 80.7 & 78.9 \\
        SARB*\cite{re8}                  & AAAI$'$22 & clip & 75.5 & 78.5 & 79.0 & 79.5 & 80.4 & 80.2 & 80.8 & 80.6 & 80.8 & 79.4 \\
        DualCoOp\cite{re12}              & NeurIPS$'$22 & clip & 78.7 & 80.9 & 81.7 & 82.0 & 82.5 & 82.7 & 82.8 & 83.0 & 83.1 & 81.9 \\
        DualCoOp++\cite{re40}             & TPAMI$'$23 & clip & 81.4 & 83.1 & 83.7 & 84.2 & 84.4 & 84.5 & 84.8 & 85.0 & 85.1 & 84.0 \\
        SCPNet\cite{re31}                & CVPR$'$23 & clip & 80.3 & 82.2 & 82.8 & 83.4 & 83.8 & 83.9 & 84.0 & 84.1 & 84.2 & 83.2 \\
        TaI-DPT\cite{re14}      & CVPR$'$23 & clip & 81.5 & 82.6 & 83.3 & 83.7 & 83.9 & 84.0 & 84.2 & 84.4 & 84.5 & 83.6 \\
        TRM-ML\cite{re30}                & ACM MM$'$24 & clip & \textbf{83.3} & \textbf{84.5} & 85.0 & 85.3 & 85.6 & 85.8 & 86.1 & 86.4 & 86.5 & 85.4 \\
        PU-MLC*\cite{re29}                & ICME$'$24 & clip & 80.2 & 83.2 & 84.4 & 85.6 & 85.9 & 86.6 & 87.0 & 87.1 & 87.5 & 85.3 \\
       T2I-PAL\cite{re16}      & TPAMI$'$25 & clip & 82.7 & 83.0 & 84.5 & 84.6 & 84.8 & 85.0 & 85.6 & 85.8 & 85.9 & 84.7 \\
        CSL*                 & Ours & clip & 82.6 & \textbf{84.5} & \textbf{85.8} & \textbf{86.6} & \textbf{87.1} & \textbf{87.5} & \textbf{87.7} & \textbf{88.0} & \textbf{88.2} & \textbf{86.4} \\
        \bottomrule
    \end{tabular}
    }
\end{table}

\textbf{Performance on VOC2007.} 
Table 4 presents the performance comparisons of different weakly supervised learning methods on VOC2007 under varying known label ratios. As illustrated in Table 4, under the same settings (i.e., using the same pre-trained backbone and input resolution), our method achieves consistently superior performance compared to ImageNet pre-trained weight-based methods. Specifically, it surpasses SST, SARB and PU-MLIC by 3.4$\%$, 3.1$\%$ and 1.8$\%$, respectively. When benchmarked against CLIP-based methods, our approach also demonstrates prominent competitive advantages. It outperforms most existing CLIP-based weakly supervised methods on VOC2007, and specifically surpasses DualCoOp, SCPNet and PU-MLC* by 1.8$\%$, 1.5$\%$ and 1.3$\%$ in average mAP, respectively. This result highlights that our method can effectively improve overall multi-label recognition accuracy, in particular, under incomplete label scenarios.
 
\begin{table}[htbp]
    \centering
    \footnotesize
    {\textbf{Table 4:} Performance comparisons (in mAP, $\%$) of different methods on VOC2007 under varying known label ratios $p$. The best performance in each comparison group is highlighted in \textbf{bold}. Note that “vanilla” refers to the image encoder pre-trained on ImageNet, whereas “clip” denotes the image encoder that adopts parameters derived from CLIP.} \\[0.5em]
    \label{Table.4}
    \resizebox{\textwidth}{!}{%
    \begin{tabular}{l l l ccccccccc c}
        \toprule
        Method & Venue & Backbone & $p{=}0.1$ & $p{=}0.2$ & $p{=}0.3$ & $p{=}0.4$ & $p{=}0.5$ & $p{=}0.6$ & $p{=}0.7$ & $p{=}0.8$ & $p{=}0.9$ & Avg \\
        \midrule
        Partial BCE\cite{re6}           & CVPR$'$19 & vanilla & 80.7 & 88.4 & 89.9 & 90.7 & 91.2 & 91.8 & 92.3 & 92.4 & 92.5 & 90.0 \\
        SSGRL\cite{re20}                 & ICCV$'$19 & vanilla & 77.7 & 87.6 & 89.9 & 90.7 & 91.4 & 91.8 & 91.9 & 92.2 & 92.2 & 89.5 \\
        P-GCN\cite{re19}                 & TPAMI$'$21 & vanilla & 83.1 & 85.9 & 87.9 & 89.2 & 90.1 & 91.3 & 92.0 & 92.7 & 93.0 & 89.5 \\
        ASL\cite{re35}                     & ICCV$'$21 & vanilla & 82.9 & 88.6 & 90.0 & 91.2 & 91.7 & 92.2 & 92.4 & 92.5 & 92.6 & 90.5 \\ 
        SST\cite{re7}                   & AAAI$'$22 & vanilla & 81.5 & 89.0 & 90.3 & 91.0 & 91.6 & 92.0 & 92.5 & 92.6 & 92.7 & 90.4 \\
        SARB\cite{re8}                   & AAAI$'$22 & vanilla & 83.5 & 88.6 & 90.7 & 91.4 & 91.9 & 92.2 & 92.6 & 92.8 & 92.9 & 90.7 \\
        PU-MLC\cite{re29}                & ICME$'$24 & vanilla & 88.0 & 90.7 & 91.9 & 92.0 & 92.4 & 92.7 & 93.0 & 93.4 & 93.5 & 92.0 \\
        CSL                  & Ours & vanilla & \textbf{89.9} & \textbf{92.4} & \textbf{93.5} & \textbf{94.0} & \textbf{94.4} & \textbf{94.6} & \textbf{94.8} & \textbf{95.2} & \textbf{95.2} & \textbf{93.8} \\
        \midrule
        DualCoOp\cite{re12}              & NeurIPS$'$22 & clip  & 90.3 & 92.2 & 92.8 & 93.3 & 93.6 & 93.9 & 94.0 & 94.1 & 94.2 & 93.2 \\
        DualCoOp++\cite{re40}            & TPAMI'23 & clip & 92.7 & 93.4 & 93.8 & 94.0 & 94.3 & 94.4 & 94.4 & 94.7 & 94.9 & 94.1 \\ 
        SCPNet\cite{re31}                & CVPR$'$23 & clip  & 91.1 & 92.8 & 93.5 & 93.6 & 93.8 & 94.0 & 94.1 & 94.2 & 94.3 & 93.5 \\
        TaI-DPT\cite{re14}      & CVPR$'$23 & clip & 93.3 & 94.6 & 94.8 & 94.9 & 95.1 & 95.0 & 95.1 & 95.3 & 95.5 & 94.8  \\ 
        PU-MLC*\cite{re29}                & ICME$'$24 & clip  & 91.3 & 92.9 & 93.3 & 93.7 & 93.8 & 94.3 & 94.5 & 94.6 & 94.8 & 93.7 \\
        TRM-ML\cite{re30}                 & ACM MM$'$24 & clip & \textbf{93.9} & 94.6 & \textbf{94.9} & \textbf{95.3} & 95.4 & 95.6 & 95.6 & 95.6 & 95.7 & \textbf{95.2} \\ 
        T2I-PAL\cite{re16}       & TPAMI$'$25 & clip & 93.7 & \textbf{94.8} & 94.8 & 94.9 & 94.9 & 95.2 & 95.5 & 95.5 & 95.5 & 95.0 \\ 
        CSL*                 & Ours & clip & 91.7 & 93.7 & 94.6 & 95.1 & \textbf{95.6} & \textbf{95.7} & \textbf{96.1} & \textbf{96.3} & \textbf{96.3} & 95.0 \\
        \bottomrule
    \end{tabular}
    }
\end{table}

\textbf{Performance on NUS-WIDE.} As exhibited in Table 5, our proposed method achieves a significant performance over existing SOTA methods on NUS-WIDE. Specifically, when compared with an ImageNet-pretrained baseline method P-GCN, our approach demonstrates superior performance with notable gains in both mAP and average mAP. When further benchmarked against CLIP-based methods, our method maintains prominent advantages, outperforming DualCoOp, TaI-DPT and T2I-PAL by 8.8$\%$, 7.7$\%$ and 7.5$\%$ in average mAP, respectively. These results fully verify the effectiveness of our proposed framework in handling the incompletely labeled setting on the NUS-WIDE dataset, especially its ability to outperform both traditional ImageNet-pretrained baselines and advanced CLIP-based approaches.

\begin{table}[htbp]
    \centering
    \footnotesize
    {\textbf{Table 5:} Performance comparisons (in mAP, $\%$) of different methods on NUS-WIDE under varying known label ratios $p$. The best performance in each comparison group is highlighted in \textbf{bold}. Note that “vanilla” refers to the image encoder pre-trained on ImageNet, whereas “clip” denotes the image encoder that adopts parameters derived from CLIP.} \\[0.5em]
    \label{Table.5}
    \resizebox{\textwidth}{!}{%
    \begin{tabular}{l l l ccccccccc c}
        \toprule
        Method & Venue & Backbone & $p{=}0.1$ & $p{=}0.2$ & $p{=}0.3$ & $p{=}0.4$ & $p{=}0.5$ & $p{=}0.6$ & $p{=}0.7$ & $p{=}0.8$ & $p{=}0.9$ & Avg \\
        \midrule
        P-GCN\cite{re19}              & TPAMI$'$21 & vanilla & 49.0 & 52.6 & 55.5 & 57.2 & 58.1 & 59.3 & 60.0 & 60.6 & 61.2 & 57.1 \\
        CSL               & Ours  & vanilla & \textbf{56.2} & \textbf{58.1} & \textbf{60.6} & \textbf{60.9} & \textbf{62.5} & \textbf{62.5}& \textbf{63.9} & \textbf{63.6} & \textbf{64.8} & \textbf{61.5} \\
        \midrule
        DualCoOp\cite{re12}           & NeurIPS$'$22 & clip  & 54.0 & 56.2 & 56.9 & 57.4 & 57.9 & 57.9 & 57.6 & 58.2 & 58.8 & 57.2 \\
        TaI-DPT\cite{re14}   & CVPR$'$23 & clip  & 56.4 & 57.9 & 57.8 & 58.1 & 58.5 & 58.8 & 58.6 & 59.1 & 59.4 & 58.3 \\
        T2I-PAL\cite{re16}    & TPAMI$'$25 & clip & 56.7 & 57.9 & 57.9 & 58.3 & 58.7 & 59.2 & 59.3 & 59.3 & 59.3 & 58.5 \\ 
        CSL*              & Ours & clip & \textbf{61.7} & \textbf{64.1} & \textbf{65.1} & \textbf{66.0} & \textbf{66.7} & \textbf{67.1} & \textbf{67.5} & \textbf{67.8} & \textbf{68.1} & \textbf{66.0} \\
        \bottomrule
    \end{tabular}
    }
\end{table}

\begin{table}[]
    \centering
    \footnotesize
    {\textbf{Table 6:} Ablation study of different components on three datasets (VOC2007, MS-COCO, NUS-WIDE) with incomplete labels. Region indicates the region score aggregation classifier; SA denotes the self-attention layer; SGFE represents the Semantic-Guided Feature Enhancement; SRFL refers to the Semantic-Related Feature Learning; CL stands for the Collaborative Learning. Performance is evaluated by using mAP ($\%$). Values in bold indicate the best performance.} \\[0.5em]
    \label{Table.6}
    \resizebox{\textwidth}{!}{%
    \begin{tabular}{c|cccccc|ccccc|c}
        \hline
        \multirow{2}{*}{Datasets} &\multicolumn{6}{c|}{{Components}}&  \multicolumn{5}{c|}{{Known Label Ratios $p$}} & \multirow{2}{*}{Avg} \\
         & {Baseline} & {Region} & {SA} & {SGFE} & {SRFL} & {CL} & $p$=0.1 & $p$=0.3 & $p$=0.5 & $p$=0.7 & $p$=0.9 & \\
        \hline
        \multirow{6}{*}{Pascal VOC} 
        & \textbf{\checkmark} &   &   &   &   &   & 79.8 & 90.8 & 92.6 & 93.1 & 93.4 & 89.9 \\
        & \textbf{\checkmark} & \textbf{\checkmark} &   &   &   &   & 82.1 & 91.9 & 93.0 & 93.6 & 94.1 & 90.9 \\
        & \textbf{\checkmark} & \textbf{\checkmark} & \textbf{\checkmark} &   &   &   & 86.5 & 91.9 & 93.0 & 94.2 & 94.7 & 92.1 \\
        & \textbf{\checkmark} & \textbf{\checkmark} & \textbf{\checkmark} & \textbf{\checkmark} &   &   & 89.6 & 93.5 & 94.2 & 94.4 & 95.0 & 93.3 \\
        &\textbf{\checkmark} & \textbf{\checkmark} & \textbf{\checkmark} & \textbf{\checkmark} & \textbf{\checkmark} &   & 89.7 & 93.4 & 94.3 & 94.8 & 95.1 & 93.5 \\
        & \textbf{\checkmark} & \textbf{\checkmark} & \textbf{\checkmark} & \textbf{\checkmark} & \textbf{\checkmark} & \textbf{\checkmark} & \textbf{89.9} & \textbf{93.5} & \textbf{94.4} & \textbf{94.8} & \textbf{95.2} & \textbf{93.6} \\
        \hline
        \multirow{6}{*}{MS-COCO} 
        & \textbf{\checkmark} &   &   &   &   &   & 72.5 & 78.0 & 80.3 & 81.5 & 82.5 & 78.9 \\
        & \textbf{\checkmark} & \textbf{\checkmark} &   &  &   &   & 74.9 & 79.5 & 81.2 & 82.2 & 83.1 & 80.2 \\
        & \textbf{\checkmark} & \textbf{\checkmark} & \textbf{\checkmark} &   &   &   & 76.5 & 81.3 & 83.2 & 84.1 & 84.7 & 82.0 \\
        & \textbf{\checkmark} & \textbf{\checkmark} & \textbf{\checkmark} & \textbf{\checkmark} &   &   & 77.3 & 81.6 & 83.5 & 84.4 & 85.0 & 82.4 \\
        & \textbf{\checkmark} & \textbf{\checkmark} & \textbf{\checkmark} & \textbf{\checkmark} & \textbf{\checkmark} &   & 78.1 & 82.2 & \textbf{83.8} & \textbf{84.7} & 85.3 & 82.8 \\
        & \textbf{\checkmark} & \textbf{\checkmark} & \textbf{\checkmark} & \textbf{\checkmark} & \textbf{\checkmark} & \textbf{\checkmark} & \textbf{78.2} & \textbf{82.3} & \textbf{83.8} & \textbf{84.7} & \textbf{85.4} & \textbf{82.9} \\
        \hline
        \multirow{6}{*}{NUS-WIDE} 
        & \textbf{\checkmark} &   &   &   &   &   & 51.9 & 57.4 & 59.9 & 61.3 & 62.3 & 58.5 \\
        & \textbf{\checkmark} & \textbf{\checkmark} &   &   &   &   & 53.7 & 58.9 & 60.9 & 62.2 & 63.0 & 59.7 \\
        & \textbf{\checkmark} & \textbf{\checkmark} & \textbf{\checkmark} &   &   &   & 54.1 & 59.5 & 61.8 & 62.9 & 63.9 & 60.5 \\
        & \textbf{\checkmark} & \textbf{\checkmark} & \textbf{\checkmark} & \textbf{\checkmark} &   &   & 55.8 & 60.2 & 62.4 & 63.5 & 64.4 & 61.3 \\
        & \textbf{\checkmark} & \textbf{\checkmark} & \textbf{\checkmark} & \textbf{\checkmark} & \textbf{\checkmark} &   & 56.1 & \textbf{60.7} & \textbf{62.7} & \textbf{63.9} & \textbf{64.8} & \textbf{61.7} \\
        & \textbf{\checkmark} & \textbf{\checkmark} & \textbf{\checkmark} & \textbf{\checkmark} & \textbf{\checkmark} & \textbf{\checkmark} & \textbf{56.2} & \textbf{60.7} & 62.5 & \textbf{63.9} & \textbf{64.8} & 61.6 \\
        \hline
        \end{tabular}%
    }
\end{table}

\subsection{Ablation Study}
In this section, to investigate the impact of different components on the performance of model  and validate the effectiveness of our proposed framework, we perform ablation studies. Specifically, we remove or modify individual components and evaluate the resulting model's performance on three benchmark datasets: VOC2007, MS-COCO and NUS-WIDE. In this paper, we adopt a baseline model consisting of ResNet-101 (pre-trained on ImageNet) and a global max-pooling classifier. We report mAP across known label ratios ranging from 0.1 to 0.9, with the average mAP presented in the last column of Table 6. As shown in Table 6, integrating our proposed components into the baseline yields consistent performance improvements over the baseline alone. These results demonstrate the effectiveness and significance of our proposed components for addressing multi-label recognition tasks with incomplete labels. Furthermore, it is worth noting that with the inclusion of label recovery, the mAP exhibits a significant increase under small value $p$, validating the effectiveness of our label recovery method.

\subsection{Visualization Analysis}
In this section, we perfom a detailed analysis on the VOC2007 dataset to assess the model’s ability in localizing discriminative regions and recovering missing labels. 

Figure 3 presents label-specific attention maps produced by our CSL method under a low labeling rate (label rate $p=0.1$). As shown in this Figure, the initial image features $F$ only coarsely highlight relevant regions and fail to achieve precise localization. By contrast, the refined semantic-aware features successfully locatize target objects with high accuracy. These results demonstrate that our proposed CSL approach not only strengthens semantic alignment through feature enhancement but also achieves precise region localization, which further contributes to improve classification performance even under extremely sparse annotations.

Figure 4 displays the label recovery results under a known label rate of 0.1. It can be observed that despite the severe label scarcity, our CSL method remains effective in localizing and recovering missing labels.

\begin{figure}[htbp]  
    \centering
    \includegraphics[width=1\textwidth]{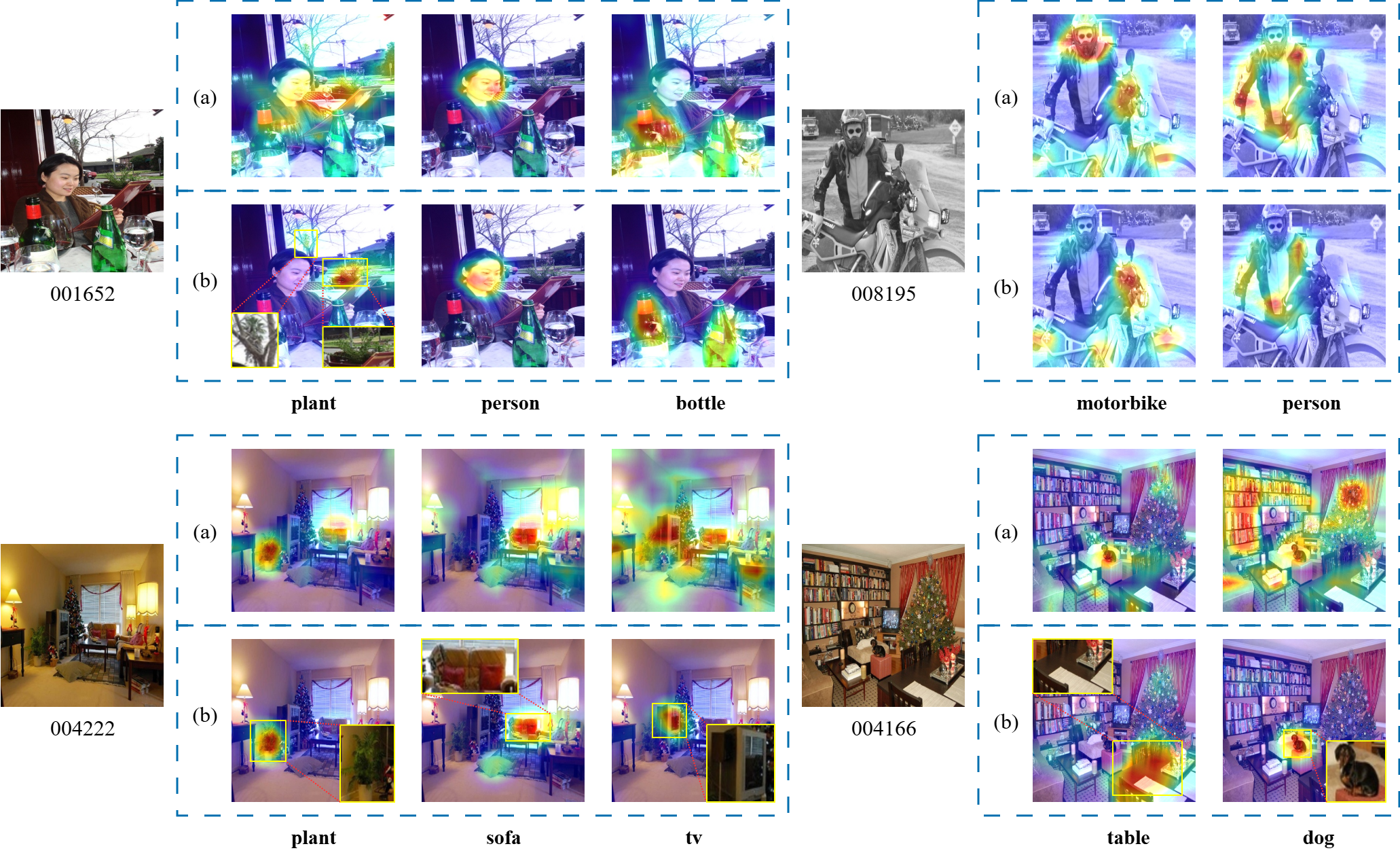}  
    \caption{Visual analysis of our CSL method. (a) Class-attention maps from image features $F$; (b) Class-attention maps from semantic-aware features $E$.}
    \label{Fig.3}
\end{figure}

\begin{figure}[htbp]  
    \centering
    \includegraphics[width=1\textwidth]{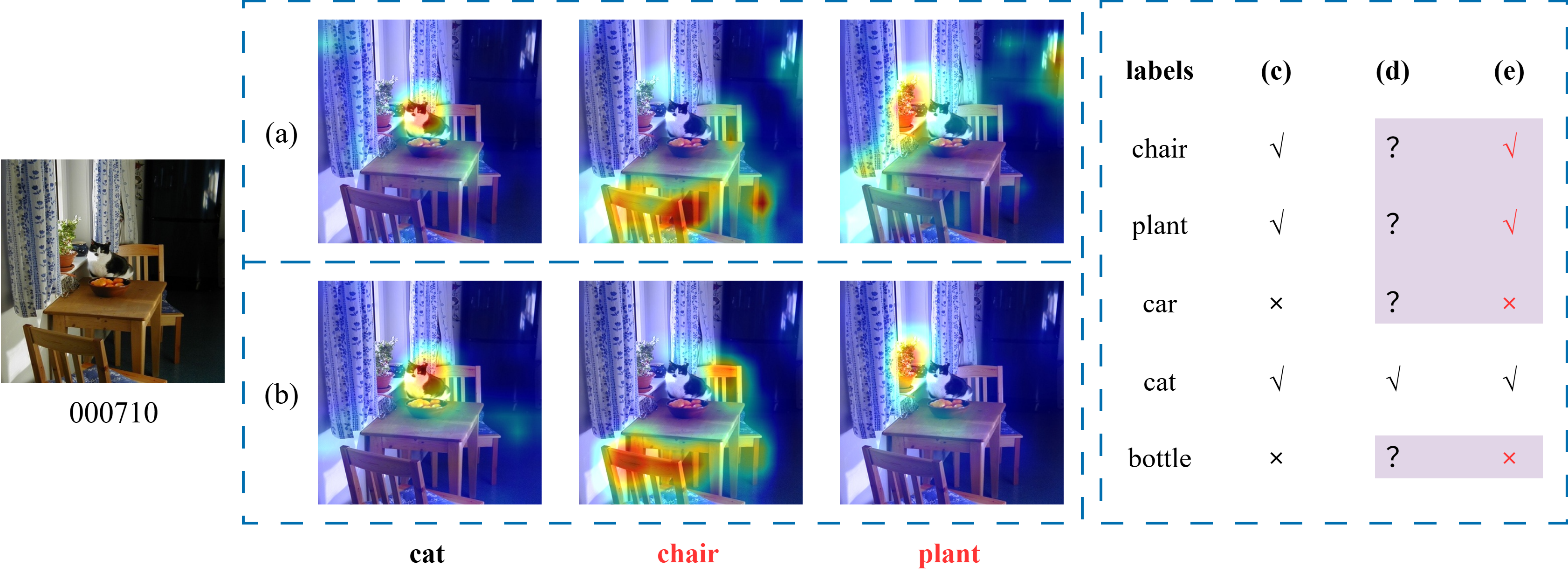}  
    \caption{Analysis of label recovery. (a) Class-attention maps from image features $F$; (b) Class-attention maps from semantic-aware features $E$; (c) Image with fully annotated labels, (d) Image with incomplete labels and (e) Image with recovered labels. In the incomplete labeled setting, some annotations (e.g., chair, plant, car and bottle) are missing.}\label{Fig.4}
\end{figure}

\section{Conclusions}

This paper proposes a collaborative learning framework for multi-label image recognition with incomplete labels, which jointly optimizes semantic-aware feature learning and missing label recovery. By effectively integrating global image features with semantic label embeddings, the proposed method not only captures inherent label correlations but also strengthens the model's ability to localize discriminative regions and interpret semantic content. Furthermore, through the joint learning of semantic-aware feature learning and label recovery, the two processes form a mutually reinforcing cycle that continuously enhances overall performance. Extensive experiments on several benchmark datasets validate the effectiveness and superiority of our proposed approach. In future work, we plan to extend this framework to more challenging scenarios, including zero-shot and few-shot multi-label image recognition.

\section*{Acknowledgements}
The work is partially funded by Natural Science Foundation of China (62041604, 62472205, 62362051), Jiangxi Provincial Natural Science Foundation (20232BAB202047, 20232BAB212013), Academic Leaders Training Program of Jiangxi Province (20232BCJ22001), Key R$\&$D Plan of Jiangxi Province (20232BBE50022).









\end{document}